\documentclass[journal,twoside]{IEEEtran}
\usepackage{amsmath,amssymb,amsfonts}
\usepackage{algorithmic}
\usepackage{accents}
\usepackage{graphicx}
\usepackage{textcomp}
\usepackage{xcolor}
\usepackage{bm}
\usepackage{pgfplots}
\usepackage[utf8]{inputenc}
\usepackage[T1]{fontenc}
\usepackage[numbers, square, comma,compress]{natbib}
\usepackage{siunitx}
\usepackage{tabularx}
\usepackage{makecell}
\usepackage{booktabs}
\usepackage{multirow}
\usepackage{orcidlink}

\pgfplotsset{compat=1.18} 
\graphicspath{{figures/}}

\begin{document}

\title{Ace! Motion Planning of Professional-Level Table Tennis Serves with a Robot Arm}

\author{Guillem Torrente$^{1}$~\orcidlink{0000-0003-2839-2774}, Guilherme Jorge Maeda$^{1}$~\orcidlink{0000-0001-9413-4787}, Divij Grover$^{2}$~\orcidlink{0000-0001-7612-4593}, Megumu Tsukamoto$^{3}$~\orcidlink{0009-0000-3668-6217}, Hamdi Sahloul$^{1}$~\orcidlink{0000-0003-3164-3028} and Peter D\"urr$^{2}$~\orcidlink{0000-0002-3840-5009},~\IEEEmembership{Senior Member,~IEEE}%
\thanks{Manuscript received: July 3, 2026; Revised: October 27, 2026; Accepted: November 7, 2026.}%
\thanks{This paper was recommended for publication by Editor Sehoon Ha upon evaluation of the Associate Editor and Reviewers' comments.}%
\thanks{$^{1}$G. J. Maeda and H. Sahloul are with Sony AI, Tokyo, Japan. G. Torrente was with Sony AI, Tokyo, Japan, when this work was done, and is now with AIRoA, Tokyo, Japan (e-mail: torrente.guillem@airoa.org).}%
\thanks{$^{2}$D. Grover and P. D\"urr are with Sony AI, Z\"urich, Switzerland.}%
\thanks{$^{3}$M. Tsukamoto is with Sony Group Corporation, Tokyo, Japan.}%
\thanks{Matches against human players followed official rules with umpires and posed no risk to the players. No private data other than videos was collected, with prior informed consent.}%
\thanks{Digital Object Identifier (DOI): see top of this page.}
}

\markboth{IEEE Robotics and Automation Letters. Preprint Version. Accepted November, 2026}
{Torrente \MakeLowercase{\textit{et al.}}: Ace! Motion Planning of Professional-Level Table Tennis Serves with a Robot Arm}

\maketitle

\begin{abstract}
Table tennis, a dynamic, compact, and popular sport, has for many years received significant attention as a robotics benchmark.
Most of the research has focused on the rally aspect---returning an incoming ball---requiring high-speed vision, agile motion planning, and tight closed-loop control.
On the other hand, getting robots to serve competitively remains a rather unexplored research problem, which in fact requires pushing physics modeling and control to the extremes.
In this work, we present the first-ever approach capable of generating official rule-compliant serves that produce ball spins of up to 550 rad/s and speeds of up to 6.7 m/s, matching and even surpassing the capabilities of elite table tennis players.
In official matches against these competitors, our serves score direct points in up to 21\% of attempts and push the win rate of robot-initiated rallies above 50\%.
The motion plans for serving are generated in simulation by combining optimization-based control and Bayesian optimization, and are executed zero-shot in open-loop on our hardware.
\end{abstract}


\section{Introduction}
Table tennis is widely regarded as a rigorous benchmark for high-speed robotics.
Characterized by its compact playing field, rapid ball exchanges, and complex aerodynamics, the sport demands systems capable of exceptional visual acuity, low-latency decision-making, and agile actuation 
\cite{ace2026, ball_stats}. 
Over the past several decades, the robotics research community has extensively utilized table tennis to advance the state of the art in these fields \cite{mulling2013learning, russell1988robot}.

\begin{figure}
    \centering
    \includegraphics[trim=0.0cm 0.0cm 0.0cm 0.0cm, clip, width=\linewidth]{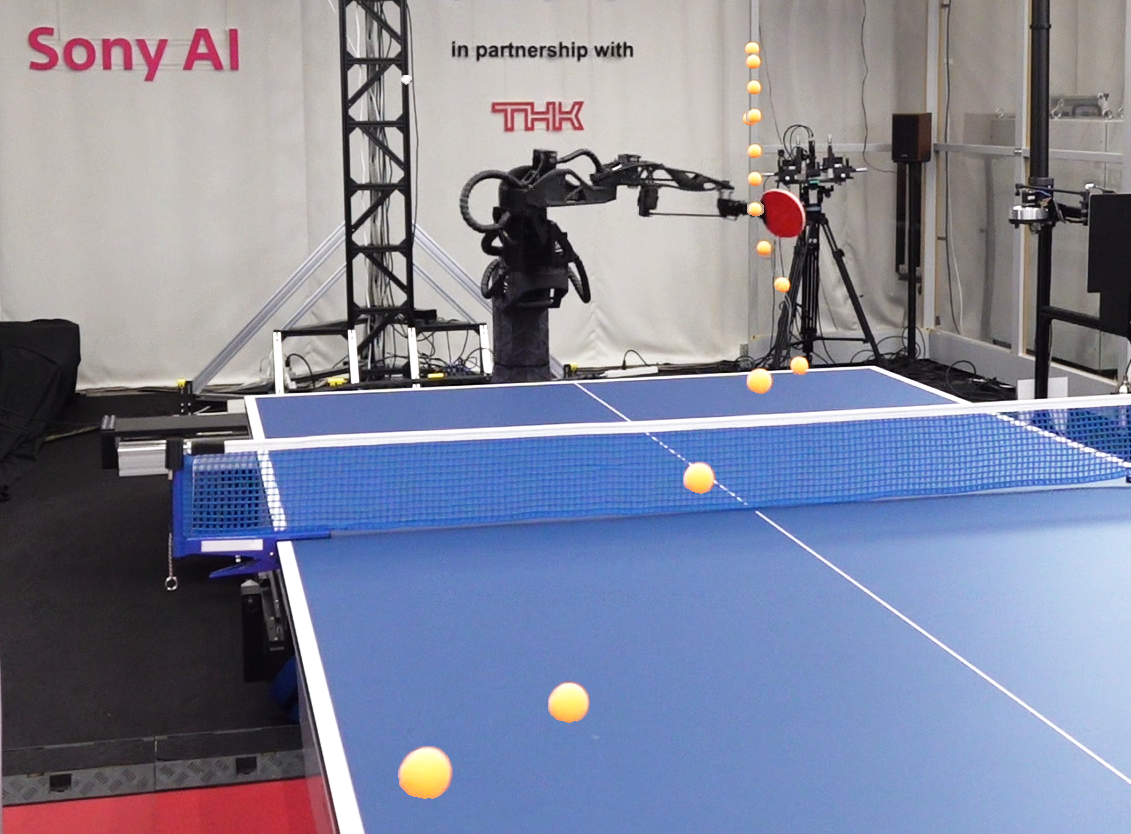}
    \caption{A high side-spin serve executed by the robot. The ball trajectory curves after each bounce due to the spin applied.
    The robot pose corresponds to approximately that at the moment of contact with the ball.}
    \label{fig:serve_main}
\end{figure}

Historically, the vast majority of research in this domain has focused on the "rally", that is, the problem of intercepting and returning an incoming ball \cite{buchler2022learning, d2024achieving}.
On the other hand, serving with a robot has received comparatively much less attention as a research problem.
This focus is understandable; the challenges of the rally map directly to fundamental and universal requirements in robotics, which necessitate robust solutions for dynamic environment perception and reactive motion planning.
Consequently, significant progress has been made in developing robot systems capable of sustaining rallies against human opponents \cite{muelling2014learning, d2023robotic, Kyohei2019ThePP}.

The serve aspect remains relatively underexplored as far as robotics research is concerned, despite being an important part of the table tennis game. 
Being the only fully controllable action during the point, elite players use the serve tactically to force certain actions on the opponent and try to secure a quick advantage \cite{Pradas:127809}.
Unlike a return, where the player leverages the momentum on the incoming ball, a competitive serve requires the generation of high-speed and/or high-spin from a spinless, free-falling ball \cite{ball_stats, Djokic2020}.
But, since the ball must bounce first on the player's own side of the table, and then on the opponent's side, there is a limit to how much energy can be applied to it.
Furthermore, the racket must intercept the ball trajectory approximately perpendicularly, making the precision requirements much higher than returning a front-incoming ball.
Athletes serve by executing complex combinations of movements involving the wrist, arm, legs and torso, generating explosive but controlled racket swings.

\paragraph*{Contributions}
In this work we present, to the best of our knowledge, the first comprehensive system for generating International Table Tennis Federation (ITTF)~\cite{ittf_statutes} compliant, elite-level serve motion plans for a robot arm (cf. Fig. \ref{fig:serve_main}).
We propose a pipeline consisting on a custom high-fidelity table tennis simulator and motion planner, that allow zero-shot transfer of a wide variety of serves in open loop, with an empirical transfer rate of $54\%$.
We maximize the diversity and difficulty of the serves by tuning the planner parameters with guided exploration using HEBO (Heteroscedastic and Evolutionary Bayesian Optimization~\cite{Cowen-Rivers2022-HEBO}), allowing us to generate challenging serves with desired characteristics.
Finally, we validate our method by playing against semi- and professional players to compare our serves against theirs in objective metrics, achieving similar or superior outcomes.

An early iteration of the presented approach was employed in a recent work, where a robot outcompeted elite table tennis players for the first time in umpire-officiated games~\cite{ace2026}.

\section{Related Work}

\subsection{Table tennis serves with robots}
Prior to the aforementioned breakthrough work in robot sports,
there have been a handful of efforts to get robot arms to perform table tennis serves, however none to our knowledge have succeeded in consistently making elite-level nor ITTF-compliant demonstrations in real hardware.

One existing approach uses kinesthetic teaching on backdrivable manipulators such as the Barrett WAM~\cite{8630019}.
Because human kinodynamics differ substantially from commercial arms, the authors use an iterative optimization algorithm to extract essential serve features into a sparse set of parameterized motions.
While this captures serve "style" and simplifies learning, it remains constrained by manual demonstrations, data quality, and hardware limitations in replicating fine wrist dexterity.


Alternatively, robots can learn serves autonomously through real-world trial-and-error~\cite{11128201}.
Using a neural network to map changes in robot motion to ball landing displacement, this system samples novel targets efficiently without extensive physical trials.
However, it optimizes solely for landing accuracy rather than high spin or velocity, and relies on an overhead ball dropper, violating ITTF toss rules.


Theoretical approaches have also addressed impact mechanics numerically~\cite{Hayakawa2016BallTrajectory}, proposing algorithms to solve for the racket orientation and velocity required to achieve target serve landing positions and spins.
Nevertheless, this work is restricted to simulations, omits manipulator kinematic and dynamic constraints, and does not account for real-world mechanical vibration or sim-to-real discrepancies.


\subsection{Control for dynamic object interception}
We employ a parameterized motion planner to control the end-effector pose and velocity at a target interception time.
Optimization-based interception for table tennis rallying was explored by~\cite{nguyen2025highspeed}, using a Non-linear Model Predictive Controller (NMPC) on a 5-DoF arm to strike balls along a fixed plane.
However, their formulation omits arm dynamics and safe post-impact deceleration.


Other work formulates the dynamic object interception task as a learning problem, training agents in simulation using either reinforcement or imitation learning~\cite{d2023robotic, catching2014}.
Furthermore, \cite{pmlr-v211-abeyruwan23a} explicitly compares the performance of a learning approach against an optimization-based controller.
Although these approaches demonstrate the feasibility of the data-driven methods, they are black box models that offer significantly less controllability regarding how the interception should be achieved, as well as robustness to out-of-distribution states.
We therefore adopt an optimization-based planner to explicitly govern impact timing and contact states.

\subsection{Parameter Optimization with HEBO}

We use HEBO to search for desired serves in the motion planner parameter space.
HEBO has been applied in several black-box optimization applications beyond standard hyperparameter tuning.
In robotics, it was used to optimize the parallel-elastic knee design parameters of the quadruped robot ANYmal~\cite{Bjelonic2023ParallelElastic}, evaluating candidates with a design-conditioned locomotion policy.
In biological sequence design, HEBO was benchmarked for combinatorial Bayesian optimization of antibody CDRH3 sequences~\cite{Khan2023AntBO}.


\section{Methodology} \label{sec:method}

\begin{figure}
    \centering
    \includegraphics[trim=0.0cm 30.0cm 0.0cm 5.0cm, clip, width=0.9\linewidth]{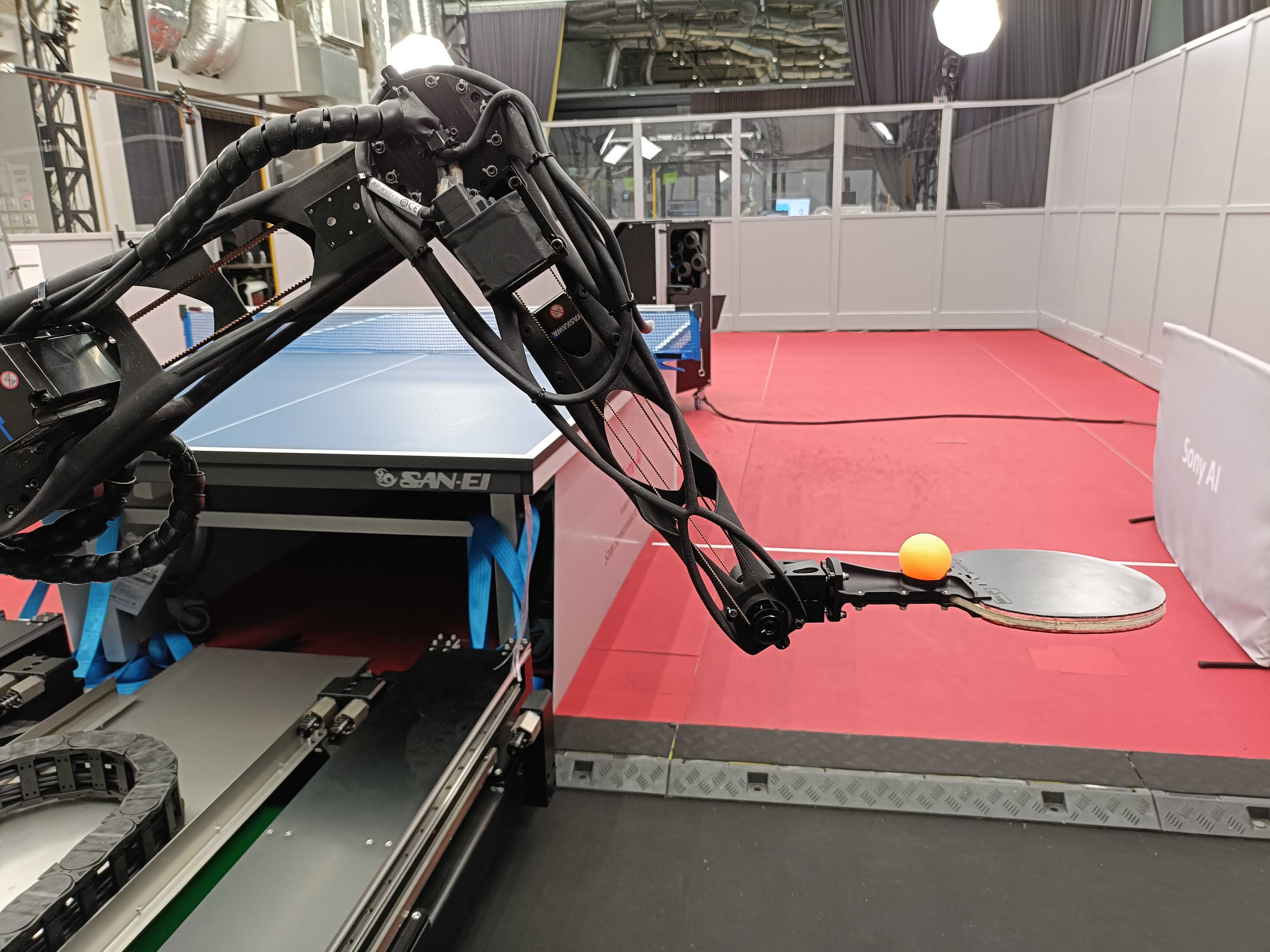}
    \caption{Our robot arm holding a ball on the ball cup, about to serve.}
    \label{fig:arm}
\end{figure}

We use a position-controlled 8-DoF robot composed of a gantry with two perpendicular prismatic axes, and an arm mounted on top with 6 revolute joints.
The 2nd, 3rd and 4th joints of the arm rotate in the same plane, compounding the speed at the End Effector (EE), where the racket is attached and whose center can reach speeds of up to $22\si{m\s}$ with a cycle time of $0.8\si{s}$.
All the actuators are synchronized using EtherCAT, providing a communication API with $2\si{ms}$ of one-way latency and a control cycle of $1\si{kHz}$~\cite{mukai2026hardwaredesigntabletennis}.
The racket also features a \emph{ball cup} in its handle, where a single ball can be placed for tossing (see Fig. \ref{fig:arm}).
With it, our robot can perform single-handed serves, a valid strategy used by some athletes with physical conditions that prevent them from doing a two hand toss, and is accepted by the ITTF rules.

In this section, we explain how we generate the serve motion plans, which are composed of \emph{preparation}, \emph{toss} and \emph{strike}.
During the \emph{preparation} phase, the ball is placed in the ball cup by a ball feeding system located in a corner of the robot court area, and the robot proceeds to move to the start of the \emph{toss} motion while keeping the racket tilt fixed to prevent the ball from falling.

\subsection{Notation, simulation setup, and robotic system}

Vector and matrix values are denoted in bold font (matrices are uppercase).
We denote the position input commands for the robot at time $t$ as $\bm{u}(t)$.
Time derivatives are denoted with the dot operator $\dot{x}$.
The robot system requires $\bm{u}(t)$, $\dot{\bm{u}}(t)$ and $\ddot{\bm{u}}(t)$ to be continuous.
The augmented robot joint state is written as $\bm{X}(t)$, an $(8\times2)$ matrix, where the first column contains the actual joint positions and the second one an auxiliary state used for more accurate numerical modeling.
We denote the 3D ball position trajectory by $\bm{p}(t)$ and its velocity by $\bm{v}(t)$, and use the subindeces $t$ and $s$ to a certain variable belonging to the \emph{toss} and \emph{strike} actions respectively.

A second-order state-space model, identified from sensor measurements, of the form $\dot{\bm{X}}(t)=\bm{f}_D(\bm{X}(t), \bm{u}(t))$ is used for simulating our robot.
It describes the linear, continuous-time dynamics of the joints, under the assumption that all axes are decoupled (due to the utilization of reduction gears).
For every joint, a continuous sub-space model is identified and later refined into a second order over-damped process model~\cite{mukai2026hardwaredesigntabletennis}.
The physics model for the ball flight considers the aerodynamic drag and Magnus forces, and for the racket and table impacts we use use instantaneous-contact models with data-based residuals on some of the components~\cite{conti2026physicsmodelssimtorealtransfer}.

The robot workspace is equipped with a state-of-the-art perception system, composed of 9 RGB cameras that triangulate the ball by detecting its color and shape, with a spatial resolution of $3\si{mm}$, and 3 event-based cameras that measure its spin with $9\%$ magnitude error.
It can also track the player's body and racket poses in real-time~\cite{ace2026}.
The origin of a right-handed coordinates system is defined at the center of the table surface, $+X$ pointing towards the human and $+Z$ towards the ceiling.

\subsection{Ball toss using motion primitives}\label{sec:toss}

Using the aforementioned perception system, we collect human demonstrations of one-arm tosses, and re-target them to the robot's kinematics via an optimization procedure \cite{maeda2016acquiring}. 
The resulting motion is a trajectory of joint commands $\bm{u}_t (t)$ that produces a ball toss when executed by the robot.
The trajectory of the racket must start with the racket pose in a "flat" state (as in Fig. \ref{fig:arm}), such that configuration can be reached from the ball feeder without dropping the ball along the way.

We define $t_{\text{{lift}}}$ as the estimated time along $\bm{u}_{t}(t)$ in which 
the ball is released from the cup.
Every toss $\bm{u}_{t} (t)$ is executed multiple times on the robot to analyze the resulting ball distributions and identify $t_{\text{{lift}}}$ empirically.
Note that, despite the executed joint commands being identical, robot vibrations and other stochastic factors such as small air flows in the room generate variability in the ball flight paths, as seen in Fig. \ref{fig:toss_spread}.

\begin{figure}
    \centering
    \resizebox{\linewidth}{!}{\input{figures/spread.pgf}}
    \caption{Spread, i.e. deviation from the expected flight path, of a ball tossed 15 times. $0.7\si{s}$ after the lift, the position uncertainty is approximately $\pm1\si{cm}$ in all directions.}
    \label{fig:toss_spread}
\end{figure}

\subsection{Parametrized motion planner for dynamic ball intersection}

\newcommand{\SetP}{\mathbb{P}}
\newcommand{\lph}{\hat{\lambda}_p}
\newcommand{\epsp}{\epsilon_p}
\newcommand{\lnh}{\hat{\lambda}_n}
\newcommand{\epsn}{\epsilon_n}
\newcommand{\lvh}{\hat{\lambda}_v}
\newcommand{\epsv}{\epsilon_v}
\newcommand{\qd}{\dot{\bm{q}}}

The striking motion $\bm{u}_s(t)$ is generated by an optimization-based motion planner that solves a non-linear hybrid problem in EE and joint spaces.
The optimization outputs a (locally) minimum-jerk control sequence $\dddot{\bm{u}}(t)$ that simultaneously satisfies initial non-stationary state constraints $(\bm{u}_{0}, \dot{\bm{u}}_{0}, \ddot{\bm{u}}_{0}, \bm{X}_0)$, a fixed resting joint state configuration $\bm{u}_{\tau_f}$ at time $\tau_f$, and EE constraints at time $\tau$, where $0 <\tau<\tau_f$.
These EE constraints define the racket position, normal axis, and velocity $\left(\bm{p}_{\tau}, \bm{n}_{\tau}, \bm{v}_{\tau}\right)$ respectively. 

To make the generated plans more precise on the real hardware, we also explicitly consider the dynamics model $\bm{f}_D$ in the optimization, such that the EE constraints are applied to the predicted robot state $\bm{X}(\tau)$, as opposed to the control state $\bm{u}(\tau)$.
This is crucial as the actual joint positions lag behind the control input due to communication and control delays, which may translate into an instantaneous racket position difference of up to a few centimeters.
We implement a custom motion planner to address all of these requirements.
For simplicity of notation, the parameters for the planning problem are aggregated into a single variable:
\begin{align}
\label{eq:xi}
\bm{\xi}_{s} \doteq \left(\bm{u}_{0}, \dot{\bm{u}}_{0}, \ddot{\bm{u}}_{0}, \bm{X}_{0}, \tau, \bm{p}_{\tau}, \bm{n}_{\tau}, \bm{v}_{\tau}, \tau_f, \bm{u}_{\tau_f}\right)
\end{align}
And the optimization problem follows:
\begin{equation}
\begin{aligned}
\label{eq:nlp}
    \min\limits_{\bm{z}} \tfrac{1}{2} \dddot{\bm{u}}^T\dddot{\bm{u}}& + \lph \epsp + \lnh \epsn + \lvh \epsv  \\
    \text{s.t. } (\bm{z}, \bm{q}_{\tau}, \qd_{\tau}) & \in \SetP(\bm{u}_{0}, \dot{\bm{u}}_{0}, \ddot{\bm{u}}_{0}, \bm{X}_{0}, \tau, \tau_f, \bm{u}_{\tau_f})  \\
    \| \bm{p}(\bm{q}_{\tau}) - \bm{p}_{\tau} \|_2 & \leq \Delta p_{\tau} + \epsp \\
    -\bm{n}_{\tau}^T \bm{n}(\bm{q}_\tau) & \leq -\cos(\Delta \phi_{\tau}) + \epsn  \\
    \| \bm{J}_{v}(\bm{q}_{\tau}) \qd_{\tau}- \bm{v}_{\tau} \|_2 & \leq \max \big\{ \Delta v_{a, \tau}, \Delta v_{r, \tau} \| \bm{v}_{\tau} \|_2\big\} + \epsv  \\
    \epsp, \epsn, \epsv & \geq 0 
\end{aligned}
\end{equation}

For speed purposes, the problem in \eqref{eq:nlp} is formulated as the optimization of a continuous chain of $n_l=32$ 
(chosen empirically to balance solver time and feasibility)
3rd degree order polynomials, that thus yields $32$ piece-wise constant jerk values $\dddot{\bm{u}}(t)$ between $[0, \tau_f]$.
The jerk solution can be integrated and sampled at any arbitrary frequency (in our case $1\si{kHz}$) to obtain $\bm{u}_s(t)$.

The decision variables $\bm{z}$ as well as the robot joint positions and velocities at time $\tau$, $(\bm{q}_\tau, \qd_\tau)$, are constrained by the abstract polytopic set~$\SetP$, which results from $\bm{f}_D$, the kinodynamic limits of the robot, and the effect of polynomial sampling. 
All of these constraints can be formulated as affine equality and inequality constraints on $\bm{z}$. 

The objective function balances trajectory smoothness and constraint satisfaction by minimizing the integrated squared jerk alongside an $\ell_1$ penalty on non-negative slack variables $(\epsp, \epsn, \epsv)$, weighted by $(\lph, \lnh, \lvh) = (2~\si{m^{-1}}, 1, 0.2~\si{s/m})$. To ensure solver feasibility, the contact constraints for position, paddle orientation, and linear velocity are softened around nominal tolerance bounds $(\Delta p_{\tau}, \Delta \phi_{\tau}, \Delta v_{a, \tau}, \Delta v_{r, \tau}) = (5~\si{mm}, \ang{0.5}, 0.01~\si{m/s}, 0.01)$. Following convergence, an \emph{a posteriori} check confirms that all slack values remain strictly negligible (i.e., the nominal tolerances were not violated).
Finally, in order to verify collision safety, a check is performed by running the motion plan through a custom collision detector based on Coal \cite{coal}.
The optimization is formulated and solved using KNITRO \cite{Byrd2006}, with typical solution times in the order of $10$ to $20~\si{ms}$.

\subsection{Optimizing serve parameters}
A serve type is defined by the ball state (spin, position, velocity, etc.) at the time of the second bounce.
After the racket impact, the ball must bounce on the table surface twice, one on each player side, and travel through the air and over or around the net.
All of these events are described using non-linear models, making the analytical calculation of desirable $\bm{\xi}_s$ very challenging.
Furthermore, only a subset of all possible $\bm{\xi}_s$ are feasible for the problem (\ref{eq:nlp}), produce collision-free trajectories, and are generate ITTF-compliant serves.

We therefore optimize $\bm{\xi}_s$ via HEBO, a surrogate-based Bayesian optimizer.
We favor HEBO over traditional GAs for two main reasons: (i) its surrogate model effectively handles the stochastic toss variations shown in Fig.~\ref{fig:toss_spread}, improving real-world transfer, and (ii) its guided exploration navigates the sparse, (up to) 10-dimensional parameter space more sample-efficiently than unguided evolutionary search.
We compare the two methods experimentally in Section~\ref{sec:results}.

Since the racket EE should intercept the ball at some time $\tau > t_{\mathrm{lift}}$, the pair $(\tau,\:\bm{p}_\tau)$ must belong to the predicted ball trajectory.
In the case that the average path is used, then the intercept point is calculated as: 
$\bm{p}_\tau = \bar{\bm{p}}_t(\tau) + \delta\bm{p}$, where $\delta \bm{p} \in \mathbb{R}^3$ is an optional small optimizable position
offset. The initial conditions in $\bm{\xi}_s$ are constrained by the toss motion $(\bm{u}_t(t), \bm{x}_t(t))$, since continuity up to acceleration must be maintained, and $\bm{u}_{\tau_f}$ is also fixed to the neutral position of the rally control policy. 
Therefore, the serve optimization contains up to ten decision variables: the hit-timing index $\tau\in\mathbb{R}^1$, the desired racket velocity at contact $\bm{v}_\tau\in\mathbb{R}^3$, the racket normal at contact parametrized by two Euler angles $\bm{n}_\tau\in\mathbb{R}^2$, the return-to-rest duration $\tau_f\in\mathbb{R}^1$ and, optionally, the position offset $\delta \bm{p} \in \mathbb{R}^3$. 
The position offset is bounded by
$[\pm 0.07, \pm 0.07, \pm 0.07]~\si{m}$, the racket velocity by
$[\pm 10, \pm 5, \pm 5]~\si{m/s}$, and the yaw-roll angles by
$[\pi/2 \pm \pi, \pm \pi]~\si{rad}$.


If the motion planner does not yield a feasible trajectory, the evaluation is assigned a fixed hard penalty of $P_{\mathrm{Fail}} = -10$ and no rollout reward is computed.
Otherwise, the produced serve is evaluated for correctness according to a sequential chain of checks, each more restrictive, until one does not pass, if any.
If the simulated serve is deemed incorrect, the rollout returns a shaping term $-10 < s_{\mathrm{leg}} < 0$.
The implemented checks and their penalties are, in order: contact with a non-racket robot body ($-5$), more than one racket--ball contact ($-4$), fewer than two table bounces ($-3$), first bounce on the wrong side of the net ($-2.5$), net collision ($-2$), second bounce on the wrong side of the net ($-1.5$), bounce too close to an edge ($-0.5$), insufficient net clearance ($-0.3$), and (optionally) failure of the robot to return to the rest configuration by the time of the second bounce ($-0.1$).

For legal serves, the reward is computed from the ball state at the second bounce and from the maximum ball height during the serve. The total reward is defined as a combination of the following terms, which can selectively be enabled or disabled depending on the design choices:
\begin{equation}
F = r_{\mathrm{pos}} + r_{\mathrm{vel}} + r_{\mathrm{top/back}}
    + r_{\mathrm{side}} + r_{\mathrm{height}} .
\label{eq:serve_reward}
\end{equation}

Here, $r_{\mathrm{pos}}$ rewards serves that land close to the specified target, whereas $r_{\mathrm{vel}}$ rewards or penalizes large rebound velocity magnitudes.
We find that penalizing velocity can be useful for encouraging short or low serves, by adding negative rewards to the XY or Z velocity components respectively.
The term $r_{\mathrm{height}}$ is inversely proportional to the maximum ball height during the serve after the first bounce, thereby encouraging low-bouncing serves and trajectories that pass close to the net. 
The generic spin terms $r_{\mathrm{top/back}}$ and $r_{\mathrm{side}}$ can be used to reward signed or unsigned spin components directly, allowing different types of spin to be selected.
The spin components are aligned with the ball travel direction, so that pure topspin corresponds to +Y spin and pure corkspin corresponds to +X spin in that frame.

Each serve candidate $\bm{\xi}_s$ is evaluated over multiple recorded toss instances to account for stochastic dynamics. By aggregating the rewards across these instances, HEBO is encouraged to select serve parameters that are robust to perturbations in the incoming ball trajectory.

\subsection{Real-world serve selection}
Finally, serves are executed on the real robot and evaluated by expert human players.
Serve instances deemed sufficiently challenging are added to the so-called \emph{competition library} for usage during official tournaments against elite and professional athletes.
The contents of the library itself also go through a process of iterative selection, albeit according to empirical observations, where the statistically worst serves are replaced by new candidates.
Over the course of multiple tournaments, the contents of the serve competition library are tracked, and serves whose rallies tend to end in opponent points are removed.

Furthermore, a heuristics-based serve selection algorithm is implemented to automatically select the next serve to use during tournaments.
This selection is based on maximizing the spin dissimilarity of consecutive serves.
Serves are clustered by spin type and magnitude, and the strategy picks the least-used serve from a cluster different from the previous one.
This keeps consecutive serves spin-distinct while minimizing repetition, limiting the opponent's ability to learn any particular serve.

\section{Results}  \label{sec:results}


\subsection{Comparison of HEBO against classic GA}
\begin{figure}
    \centering
    \includegraphics[width=0.95\linewidth]{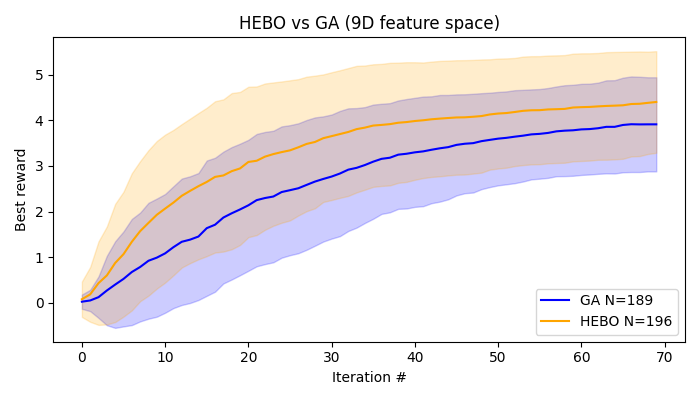}
    \caption{Comparison of convergence of HEBO against GA parameter search over 70 iteration-long runs with equivalent settings (200 samples for each). Thick lines represent mean values, shaded areas represent standard deviations. HEBO outperforms the GA as the final average reward is $15\%$ higher.}
    \label{fig:hebo_vs_ga}
\end{figure}

First, we study the ability of HEBO to find competitive serves faster than standard GA.
For the same target serve characteristics, we initialize 200 instances of the HEBO and a GA implemented with PyGAD~\cite{gad2023pygad}.
The number of search parameters is 9, as we leave the choice of $\tau_f$ up to the motion planner.
Both algorithms run for 70 iterations, with a population size of 20, and with the initial population being uniformly sampled.
We use the default parameters for the GA algorithm, except for the maximum number of simultaneous mutations, which we set to 2, and the number of surviving individuals, which we set to 6.
Since the GA exploration is random, we choose to slightly restrict it so that local optima are reached more easily once an \emph{feasible} serve is found.
On the other hand, HEBO has the opposite tendency of exploiting the known areas of high reward and missing out on better serves, so instead we encourage exploration by randomly sampling 25\% of the candidates every iteration.
The convergence results are illustrated in Fig.~\ref{fig:hebo_vs_ga}, where HEBO reaches higher reward values earlier.
We also observe that while 4 out of the 200 runs of HEBO did not find any legal serve, that number rose to 11 for the GA.
We believe this is due to the fact that even if the GA finds feasible but invalid serves, it does not have a guided way to explore the local parameter space to find those serves that satisfy all rules. 

\begin{figure*}
    \centering
    \includegraphics[width=\textwidth]{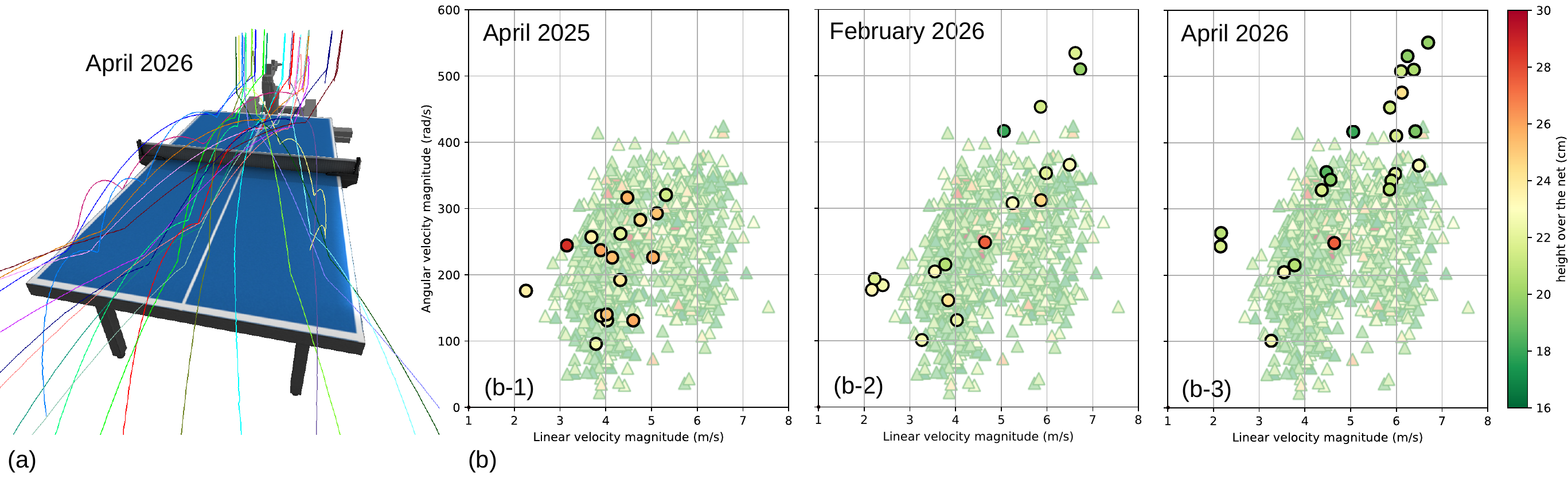}
    \caption{
(a) Real ball trajectories of serves used during the match against the Paris 2024 Olympic silver medalist Miu Hirano.
(b) Real instances of ball velocity-spin data pairs from serves performed by our robot (thick dots) and human athletes (faded triangles). The ball state is reported as it crosses the net plane, and the color gradient corresponds to the passing height. The robot data is shown as it evolved across 3 different competitions. The human data is displayed for comparison, and includes all 1243 serves performed by professional players against our robot during the same period of time.
Starting from February 2026, the spin magnitude of robot serves exceeded those of the professional players.
    }
    \label{fig:serve_metrics}
\end{figure*}

\subsection{Serve proficiency}
We claim that the serves generated with our approach are capable of matching and, according to numerical metrics of spin and velocity, outperform those of world-class players.
To empirically assess the serves, our robot plays against elite- and professional-level athletes of the Japanese leagues in tournaments officiated by licensed umpires.
Because of the players' limited availability to come to our laboratory and the cost involved in coordinating these experiments, the presented method is improved iteratively after analyzing the results of every session.

Next, we detail the sequential improvements made to the serve generation pipeline together with performance metrics during player matches, enumerated in Table \ref{tab:serve_metrics}.
The first row indicates the baseline method 
as outlined in the original work~\cite{ace2026}, and 
used during the April 2025 tournament, and the last row corresponds to the final method as described in Section~\ref{sec:method}.
For every iteration (row) of the serve motion planning pipeline, we provide an estimate of the probability of our robot winning a point, both when serving and when receiving. 
We use the Binomial 95\% Wilson Confidence Intervals (WCI) for calculating a confidence range based on the limited data available.
Also, we report the number of direct serve points (Aces) scored by our robot.
Note that, as our pipeline and the generated serves improve, also does the probability of winning the point by serving, while the probability of winning the point by receiving remains relatively unchanged.
That is, despite the fact that later experiments were mostly conducted against professional players, whose level is higher than those of elite players, who we mostly invited during earlier sessions.

On the first iteration, the serve simulation relies only on the mean trajectory $\bar{\bm{p}}_t(t)$ for simulating the ball toss. 
Starting in November 2025, instead of using $\bar{\bm{p}}_t(t)$, for every serve training episode, we sample multiple ball toss trajectories from the recordings and compute the serve average reward over all.
This helps make our serves more robust, which we verify empirically as the reliability of open-loop execution improves.
Ball tosses from both sides of the robot are also introduced, increasing variability and making adaptation more difficult for players.

The method described in Sec. \ref{sec:method} is fully deployed during the December 2025 tournament, with HEBO replacing the classic GA optimizer used until that point. 
Then, the derivative limits used by the motion planner are increased such that joint acceleration and jerk limits are higher, as we found that the torque usage for every joint was still within the saturation levels.
This allows to generate a set of motion plans producing increased linear and angular ball velocities, which can be clearly observed in Fig. \ref{fig:serve_metrics}(b).
This figure shows the state of the robot's library in three different stages, compared against professional human serve data.

For the tournaments in March and April 2026, we also strategically prune and replace serves whose historical performance shows a tendency to lead to opponent's points during the rally.
This way of modulating the library leads to an increase in the robot's direct serve points---a situation in which the opponent fails to return the ball legally, also sometimes called ``Aces''---as the average speed of serves in the library also increases. 

According to pre-existing analyses of elite table tennis matches, points won while serving are approximately 52.78\% and 53.28\% for professional male and female players, respectively~\cite{gomez2017analysis}. 
Therefore, the ability of our robot to maintain slightly higher winning probabilities while serving (51.5 to 58.7\% in April 2026) suggests that the serves achieve comparable effectiveness to top professional human players, even against notable players such as the two-time Olympian and silver medalist Miu Hirano.

\begin{table*}
\centering
\caption{Serve statistics against elite and professional players}
\label{tab:serve_metrics}
\footnotesize
\setlength{\tabcolsep}{3pt}
\renewcommand{\arraystretch}{1.15}

\begin{tabularx}{\textwidth}{
l@{\hspace{20pt}}
>{\raggedright\arraybackslash}X
>{\raggedright\arraybackslash}X
c@{\hspace{10pt}} c@{\hspace{10pt}} c c c
}

\toprule

\makecell{\textbf{Date}} & \textbf{Participants} & \textbf{Feature} &
\makecell{\textbf{Serves in}\\\textbf{library}\\(right, left)} &
\makecell{\textbf{Total robot}\\ \textbf{serves during}\\ \textbf{the game}} &
\multicolumn{2}{c}{\makecell{\textbf{Cumulative} \\
\textbf{win probability}\\
\textbf{95\% WCI} [\%]}} &
\makecell{\textbf{Serve Points} \\ ``Aces'' \\ (\%) } \\

\cmidrule(lr){6-7}

& & & & &
\makecell{receiving} &
\makecell{serving} &
\\

\midrule

Apr. 2025 &
\makecell[l]{3 elite females\\
3 elite males\\
Pro male: Kakeru Sone\\
Pro female: Minami Ando} &
\makecell[l]{Baseline} &
\makecell{15\\(15, 0)} &
177 &
44.7--59.2 &
40.8--55.3 &
\makecell{20\\ (11\%)} \\ 

\midrule

Nov. 2025 &
\makecell[l]{
3 elite females\\
2 elite males} &
\makecell[l]{Training on tossing distributions\\Serving from both sides} &
\makecell{14\\(10, 4)} &
102 &
47.6--59.3 &
45.1--56.7 &
\makecell{9\\(9\%)} \\ 

\midrule

Dec. 2025 &
\makecell[l]{
2 elite females\\
3 elite males\\
Pro male: Kazuhiro Yoshimura\\
Pro female: Mayuka Taira} &
HEBO as optimizer &
\makecell{16\\(11, 5)} &
171 &
46.7--55.9 &
48.1--57.2 &
\makecell{14\\(8\%)} \\ 

\midrule

Feb. 2026 &
\makecell[l]{2 elite males\\
Pro female: Maki Shiomi} &
Dynamic limit expansion &
\makecell{17\\(12, 5)} &
125 &
46.9--55.0 &
49.1--57.3 &
\makecell{2\\(1.6\%)} \\ 

\midrule

Mar. 2026 &
\makecell[l]{
Pro female: Miyuu Kihara\\
Pro male: Tonin Ryuzaki\\
Pro male: Fumiya Igarashi} &
\makecell[l]{Priority to side and\\topspin serves} &
\makecell{21\\(13, 8)} &
109 &
48.6--56.1 &
50.8--58.2 &
\makecell{23\\(21\%)} \\ 

\midrule

Apr. 2026 &
\makecell[l]{Pro female: Miu Hirano} &
--- &
\makecell{22\\(12, 10)} &
40 &
49.9--57.1 &
51.5--58.7 &
\makecell{8\\(20\%)} \\ 

\bottomrule
\end{tabularx}
\end{table*}

\subsection{Task-specific serve generation example}
In this final experiment, we perform an analysis of the sim-to-real transfer of our pipeline without human fine-tuning in the real world.
We define 4 types of serving tasks and generate $25$ motion plans for each.
These are: \emph{Aiming}: the ball should land on one specific table corner, \emph{Topspin \& velocity}: maximize the ball topspin and velocity magnitude in the x-y plane, \emph{Backspin}: maximize the ball backspin, \emph{Sidespin}: maximize the ball sidespin (in any direction).
The reward weights of the HEBO algorithm are adjusted according to these objectives, and the algorithm is run $25$ times.
Besides the main task reward weights, we also add the low-height reward weight for all training runs as it helps produce human-like serves.
Other constant hyperparameters include a population size of 20 and maximum number of iterations of 100. 
Under this setup, a serve can be trained in about $10\si{min}$ in a single thread using a Desktop PC with an Intel i9-12900KF CPU.
The generated serves are finally run on the robot in open-loop.

Table \ref{tab:serve_sim_to_real} reports the statistics of the serves in simulation and in the real hardware.
Indeed, not all of the plans transferred correctly to the real world due to still existent sim-to-real gaps.
The real world statistics are thus based on the subset of serves that were successfully deployed.
Topspin and backspin are reported in the same column since the difference between the two is in their sign (negative for backspin).

The results suggest that there is a strong correlation between the serve property encouraged through the reward function and the corresponding physical metric (in bold font), confirming our ability to control the type of serve generated via HEBO exploration.
For example, when topspin is rewarded, the average value becomes significantly larger for both Sim and Real setups ($218$ to $263 \si{rad/s}$) but the sign is flipped when backspin is encouraged instead ($-131$ to $-164 \si{rad/s}$).
Since for the sidespin task we do not specify a preferred direction, the obtained mean value is relatively small, but the standard deviation is much larger compared to that of the other tasks, as both positive and negative sidespins are produced.

\begin{figure}
    \centering
    \includegraphics[trim=0.5cm 2.5cm 0.5cm 0.0cm, clip, width=\linewidth]{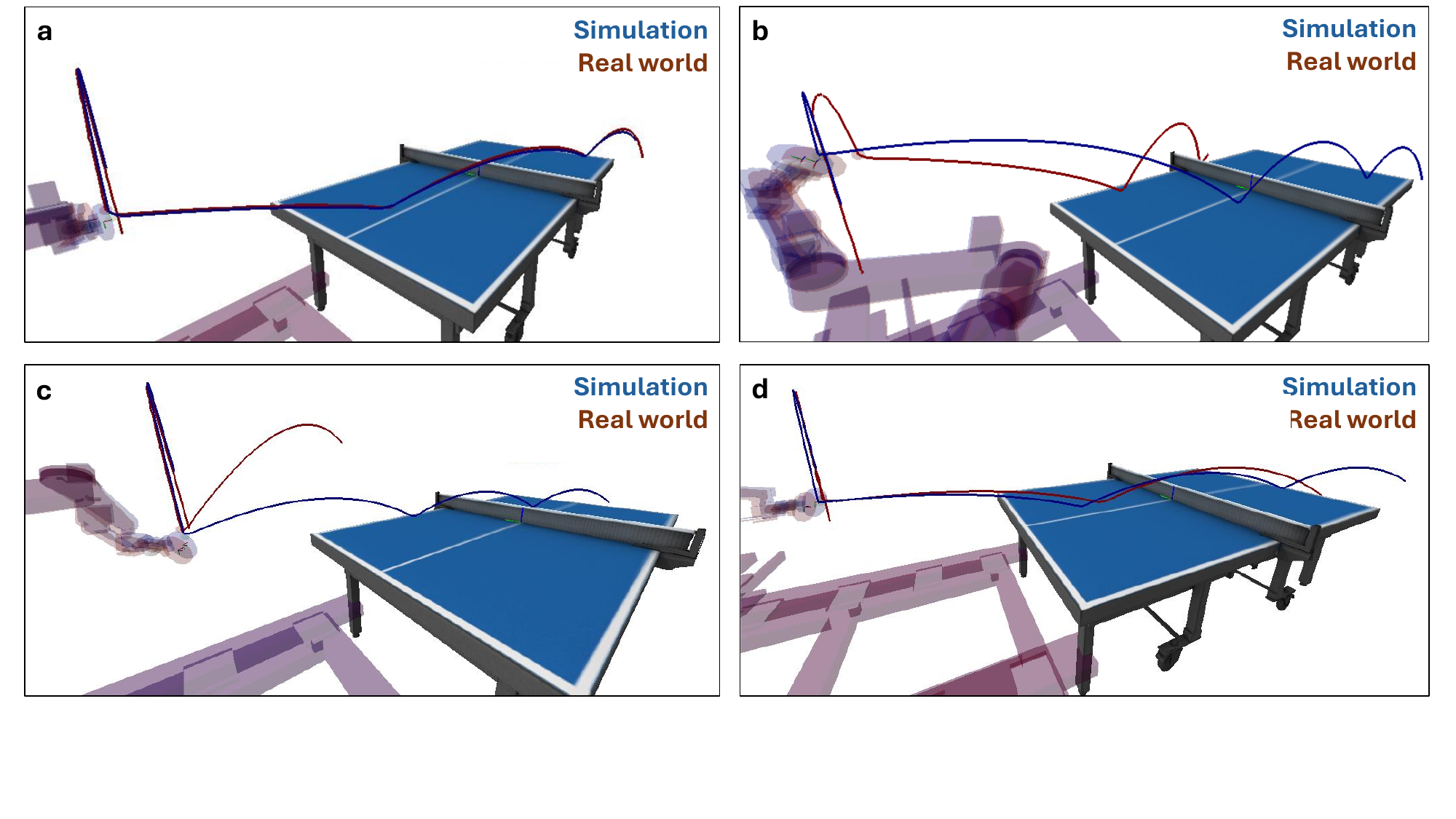}
    \caption{Comparison of four serve recordings on the hardware (red) against the simulated rollouts (blue). \textbf{a}: the real data matches the simulated one and the serve is transferred correctly. \textbf{b}: failed transfer due to a toss OOD event. \textbf{c}: failed transfer due to a calibration mismatch, causing a racket edge hit. \textbf{d}: failed transfer due to contact physics inaccuracies, causing the ball to overshoot during the second bounce.}
    \label{fig:real_vs_sim}
\end{figure}

Finally, we study the serves that did not transfer to the real world.
Qualitatively, the failure effects can be classified into 5 categories: the racket misses the ball after the toss, and the ball misses the table, or bounces at the incorrect side of it, for both bounces.
In Table \ref{tab:failure_modes}, we assign the missed serves to these categories.
For the majority of cases, the first bounce is missed, followed by the second bounce also being missed, while the rest of events are limited in number.
Then, we contrast the recordings from the robot and camera sensors against the corresponding simulated rollouts to understand the failures causes, which we divide into 3 categories and exemplify in Fig. \ref{fig:real_vs_sim}: the ball toss being Out Of Distribution (OOD), calibration-related gaps that cause the ball hit to be off (i.e. on the racket edge) or completely missed, and simulation and contact physics-related inaccuracies.
The three are roughly equally represented in the dataset.

Visual exploration of toss OOD events, which is shown in Fig. \ref{fig:real_vs_sim}\textbf{b}, reveals that the ball starts deviating from the toss distribution mid-flight, as opposed to the lift-off instant, despite no apparent contact with the racket.
We hypothesize that the racket may be creating some wind turbulence due to the proximity between the ball cup and the blade.
This failure mode is prevalent in the aiming task, where different robot serves have similar motion profiles.
For the calibration-related category (Fig. \ref{fig:real_vs_sim}\textbf{c}), issues likely come from combinations of small inaccuracies in our pipeline, such as the robot DH calibration or its dynamics model, but can also be caused by link deflection, or simply toss stochasticity.
For the physics modeling (Fig. \ref{fig:real_vs_sim}\textbf{d}), we can clearly see the ball increasingly deviating from the simulated path after contacting the racket or table, pointing to an inaccuracy in our models.

\begin{table*}[]
\centering
\caption{Serve statistics in simulation and in the real world for different task settings}
\label{tab:serve_sim_to_real}
\renewcommand{\arraystretch}{1.15}
\begin{tabular}{cc|c|ccc|ccc}
\multirow{2}{*}{\textbf{Task}}                                                       & \multirow{2}{*}{\textbf{Environment}} & \begin{tabular}[c]{@{}c@{}}\textbf{Aim error}\\ {[}mm{]}\end{tabular} & \multicolumn{3}{c|}{\begin{tabular}[c]{@{}c@{}}\textbf{Velocity}\\ {[}m/s{]}\end{tabular}} & \multicolumn{3}{c}{\begin{tabular}[c]{@{}c@{}}\textbf{Spin}\\ {[}rad/s{]}\end{tabular}} \\
                                                                            &                      & x-y norm                                                     & x                         & y                         & z                         & cork                    & top (+) / back (-)                    & side                   \\ \hline
\multirow{2}{*}{\textbf{Aiming}}                                                     & Sim $(N=25)$                 &  $\bm{166\pm285}$                                                            &  $2.6\pm1.2$                         & $-0.9\pm0.4$                          &   $2.3\pm0.3$                        &  $26.9\pm41.6$                       & $60.0\pm129.0$                            & $-12.9\pm124.6$                       \\
                                                                            & Real $(N=15)$                &  $\bm{299\pm313}$                                                            & $2.8\pm1.0$                          & $-1.0\pm0.4$                           & $2.3\pm0.3$                           &  $22.9\pm41.5$                       & $56.7\pm158.3$                             & $3.5\pm111.8$                        \\ \hline
\multirow{2}{*}{\begin{tabular}[c]{@{}c@{}}\textbf{Topspin}\\ \textbf{velocity}\end{tabular}} & Sim $(N=25)$                 &  N/A                                                            & $\bm{4.5\pm1.3}$                          & $\bm{-0.7\pm0.8}$                           & $2.0\pm0.1$                           & $33.8\pm44.2$                         & $\bm{218.0\pm75.4}$                             & $38.2\pm143.8$                        \\
                                                                            & Real $(N=14)$                & N/A            & $\bm{4.4\pm1.1}$                                                              & $\bm{-0.7\pm0.6}$                           & $2.0\pm0.1$                           & $48.8\pm54.3$                           & $\bm{263.4\pm70.0}$                         & $100.5\pm131.8$                        \\ \hline
\multirow{2}{*}{\textbf{Backspin}}                                                   & Sim $(N=25)$                 & N/A                                                             & $2.7\pm0.9$                           & $-0.9\pm0.5$                           & $1.7\pm0.1$                           & $-68.9\pm70.5$                        & $\bm{-164.7\pm128.0}$                             & $19.1\pm113.3$                        \\
                                                                            & Real $(N=12)$                 & N/A                                                             & $3.3\pm0.8$                          & $-1.1\pm0.5$                           & $1.7\pm0.2$                           & $-69.7\pm83.5$                         & $\bm{-131.8\pm176.9}$                             & $29.2\pm72.3$                        \\ \hline
\multirow{2}{*}{\textbf{Sidespin}}                                                   & Sim $(N=25)$                 & N/A                                                              & $3.5\pm0.8$                           & $-0.4\pm0.8$                           & $1.9\pm0.1$                           & $13.2\pm 44.1$                         & $124.4\pm108.0$                             & $\bm{84.3\pm243.8}$                        \\
                                                                            & Real $(N=13)$                 & N/A                                                              & $3.7\pm1.0$                           & $-0.7\pm0.7$                           & $2.0\pm0.2$                          & $36.7\pm35.2$                        & $139.7\pm79.3$                             & $\bm{-13.1\pm255.2}$                       
\end{tabular}
\end{table*}

\begin{table}[]
\centering
\caption{Failure modes and reasons of missed serves from Table \ref{tab:serve_sim_to_real}}
\label{tab:failure_modes}
\renewcommand{\arraystretch}{1.15}
\setlength{\tabcolsep}{3.5pt} 
\begin{tabular}{c|ccccc|ccc}
                                                                  \multirow{2}{*}{\textbf{Task}}   & \textbf{Racket} & \multicolumn{2}{c}{\textbf{1\textsuperscript{st} bnc.}} & \multicolumn{2}{c|}{\textbf{2\textsuperscript{nd} bnc.}} & \textbf{Toss} & \multirow{2}{*}{\textbf{Calib.}} & \textbf{Physics} \\
                                                       & \textbf{miss}   & \textbf{miss}       & \textbf{bad}       & \textbf{miss}       & \textbf{bad}  & \textbf{OOD} & & \textbf{\&Sim.}  \\ \hline
\textbf{Aiming}                                                              & 0                & 6                    &  1                  &  3                   &  0    & 8 & 1 & 1             \\
\begin{tabular}[c]{@{}c@{}}\textbf{Topspin}\\ \textbf{velocity}\end{tabular}          &  1               &  5                   &  0                  &  5                  &  0 & 0 & 6 & 5                 \\
\textbf{Backspin}                                                            &  3               &  7                   &  1                  &  0                   &  2  & 5 & 4 & 4                \\
\textbf{Sidespin}                                                            &  0               &  7                   &   0                 &  5                   &  0 & 3 & 2 & 7    \\ \hline
\textbf{Total} & 4 & 25 & 2 & 13 & 2 & 16 & 13 & 17
\end{tabular}
\end{table}

\section{Conclusions and Future work}
We have presented, to our knowledge, the first framework capable of generating professional-level, ITTF-compliant serve motion plans for a robot arm.
When evaluated against elite- and professional-level players, our serves showcased comparable or even higher amounts of velocity and especially spin to those performed by these players.
This method was one of the key components that, also for the first time in history, allowed a robot to win in Best-of-3 matches against professional table tennis players.

Our method consists of a parametrized hybrid end-effector and joint-space optimization-based motion planner, whose parameters are found in simulation by a Bayesian optimizer.
In our case, HEBO is used to find the parameters defining the racket state when intercepting the table tennis ball to perform serves with user-defined characteristics.
We expect this approach to be generalizable to other motion planning problems where the main challenge is to find a large diversity of motion profiles subject to secondary quantifiable tasks with local smoothness properties.

Possible future work could focus on two aspects: (i) reducing the transfer failure rate that hinders the efficiency of our method.
This could be achieved by a combination of improved physics modeling (including ball-racket contact, ball toss, robot dynamics, robot deflection and robot vibration), controlling more carefully the racket orientation while swinging to reduce air disturbances, as well as enabling automatic fine-tuning of motion plans, with HEBO or other black-box optimizers, in the hardware system.
And (ii), tightly coupling the serve development to the rally controller, to develop more long-sighted strategies for scoring points.

\section*{Acknowledgements}
The authors thank Stefan Richter for his contributions to the development of the motion planner, and all the players and umpires who participated in the experiments.

\bibliographystyle{IEEEtran}
\bibliography{bibliography}

\end{document}